\documentclass[11pt]{article}
\usepackage{eacl2014}
\usepackage{times}
\usepackage{url}
\usepackage{latexsym} 
\usepackage{multirow}

\title{Experiments to Improve Named Entity Recognition on Turkish Tweets}

\author{Dilek K\"u\c{c}\"uk \and Ralf Steinberger\\
  European Commission, Joint Research Centre \\
  Via E. Fermi 2749\\
  21027 Ispra (VA), Italy \\
  {\tt firstname.lastname@jrc.ec.europa.eu}\\}

\date{}

\begin{document}
\maketitle
\begin{abstract}
Social media texts are significant information sources for several application areas including trend analysis, event monitoring, and opinion mining. Unfortunately, existing solutions for tasks such as named entity recognition that perform well on formal texts usually perform poorly when applied to social media texts. In this paper, we report on experiments that have the purpose of improving named entity recognition on Turkish tweets, using two different annotated data sets. In these experiments, starting with a baseline named entity recognition system, we adapt its recognition rules and resources to better fit Twitter language by relaxing its capitalization constraint and by diacritics-based expansion of its lexical resources, and we employ a simplistic normalization scheme on tweets to observe the effects of these on the overall named entity recognition performance on Turkish tweets. The evaluation results of the system with these different settings are provided with discussions of these results.
\end{abstract}

\section{Introduction}\label{sec:intro}
Analysis of social media texts, particularly microblog texts like tweets, has attracted recent attention due to significance of the contained information for diverse application areas like trend analysis, event monitoring, and opinion mining. Tools for well-studied problems like named entity recognition (NER) are usually employed as components within these social media analysis applications. For instance, in \cite{Abel2011}, named entities extracted from tweets are used to determine trending topics for user modeling within the context of personalized recommender systems and in \cite{Ritter2012}, named entities in tweets are used to complement the events extracted by an open domain event extraction system for Twitter. However, existing NER solutions for well-formed text types like news articles are reported to suffer from considerable performance degradations when they are ported to social media texts, mainly due to the peculiarities of this latter text type \cite{Ritter2011}.

In this paper, we report on our NER experiments on Turkish tweets in order to determine facilitating and impeding factors during the development of a NER system for Turkish tweets which can be used in social media analysis applications. We carry out these experiments on two tweet data sets annotated with named entities. After the initial evaluation results of a rule-based NER system \cite{Kucuk2009} on these data sets, we gradually present the performance results achieved by the extended versions of the system together with discussions of these results. For these experiments, we first perform two system adaptations, i.e., relaxing the capitalization constraint of the system and diacritics-based expansion of the system's lexical resources. Next, we incorporate a simplistic tweet normalization scheme into the NER procedure. After the evaluation of these extensions, we provide discussions on the plausible features of a NER system tailored to Turkish tweets.

The rest of the paper is organized as follows: In Section \ref{sec:related}, we review the literature on NER on tweets and NER on Turkish texts. In Section \ref{sec:tests}, we present our NER experiments on Turkish tweets. Directions of future work are outlined in Section \ref{sec:future} and finally Section \ref{sec:conc} concludes the paper.

\section{Related Work}\label{sec:related}
There are several recent studies presenting approaches for NER on microblog texts, especially on tweets in English. Among these studies, in \cite{Ritter2011}, a NER system tailored to tweets, called T-NER, is presented which employs Conditional Random Fields (CRF) for named entity segmentation and labelled topic modelling for subsequent classification, using Freebase dictionaries. A hybrid approach to NER on tweets is presented in \cite{Liu2011} where k-Nearest Neighbor and CRF based classifiers are sequentially applied. In \cite{Liu2012}, a factor graph based approach is proposed that jointly performs NER and named entity normalization on tweets. An unsupervised approach that performs only named entity extraction on tweets using resources like Wikipedia is described in \cite{Li2012}. A clustering-based approach for NER on microtexts is presented in \cite{Jung2012}, a lightweight filter based approach for NER on tweets is described in \cite{Oliveira2013}, and a series of NER experiments on targeted tweets in Polish is presented in \cite{Piskorski2013}. Finally, an adaptation of the ANNIE component of GATE framework to microblog texts, called TwitIE, is described in \cite{Bontcheva2013}.

Considering NER research on Turkish texts, various approaches have been employed so far including those based on using Hidden Markov Models (HMM) \cite{Tur2003}, on manually engineered recognition rules \cite{Kucuk2009,Kucuk2012}, on rule learning \cite{Tatar2011}, and on CRFs \cite{Yeniterzi2011,Seker2012}. All of these approaches have been proposed for news texts and the CRF-based approach \cite{Seker2012} is reported to outperform the previous proposals with a balanced F-Measure of about 91\%.

To the best of our knowledge, there are only two studies on NER from Turkish tweets. In \cite{Eryigit2013}, the CRF-based NER system \cite{Seker2012} is evaluated on informal text types and is reported to achieve an F-Measure of 19\% on tweets. In \cite{Kucuk2014}, a tweet data set in Turkish annotated with named entities is presented. The adaptation of a multilingual rule-based NER system \cite{Pouliquen2009} to Turkish, which achieves an F-Measure of about 61\% on a news article data set, gets an F-Measure of 37\% on this tweet data set, and after extending the resources of the NER system with frequently appearing person and organization names in Turkish news articles, the corresponding scores increase to about 69\% and 43\%, respectively \cite{Kucuk2014}.

\begin{table}[h]
  \small
  \centering
  \caption{NE Statistics on the Data Sets.}
    \begin{tabular}{|l|c|c|}
    \hline
    \multicolumn{1}{|c|}{\textbf{}} & \multicolumn{2}{|c|}{\textbf{Frequency in}} \\
    \cline{2-3}
    \textbf{NE Type~~~} & \textbf{~Tweet Set-1~} & \textbf{~Tweet Set-2~} \\ \hline
    Person  & 457 & 774 \\ \hline
    Location & 282 & 191 \\ \hline
    Organization & 241 & 409 \\ \hline
    All PLOs & 980 & 1,374 \\ \hline
    Date & 201 & 342 \\ \hline
    Time & 5 & 25 \\ \hline
    Money & 16 & 13 \\ \hline
    Percent & 9 & 3 \\ \hline
    All NEs & 1,211 & 1,757 \\ \hline
    \end{tabular}
  \label{tab:stat}
\end{table}

\section{Named Entity Recognition Experiments}\label{sec:tests}
The NER experiments are performed using the rule-based NER system \cite{Kucuk2009} which makes use of a set of lexical resources, i.e., lists of person\slash location\slash organization names (henceforth referred to as PLOs), and patterns for the extraction of named entities (NEs) of type PLOs and time\slash date\slash money\slash percent expressions \cite{Kucuk2009}. The system is proposed for news articles which is a considerably well-formed text type usually with proper capitalization of the initial letters of PLOs and separation of these PLOs from their suffixes with apostrophes\footnote{An example inflected named entity of location name type (a city name) in Turkish which takes the dative case suffix ($-ya$) is $Ankara$'$ya$ (meaning $to~Ankara$) where the initial letter of the named entity is properly capitalized and the case suffix is accordingly separated from the entity with an apostrophe.}. Yet, as even such well-formed texts may be lacking these important indicators of PLOs, the system can be configured to make use of the capitalization clue or not, and it includes a simplistic morphological analyzer to check the suffixes at the end of PLO candidates and thereby validate these candidates \cite{Kucuk2009}.

This NER system achieves a balanced F-Measure of 78.7\% (without giving any credit to partial extractions) on a news article data set of about 20K tokens obtained from the METU Turkish corpus \cite{Say2002} where the annotated form of this data set includes a total of 1,613 NEs. Within the course of the current study, we have evaluated this system on two tweet data sets in Turkish where statistical information about these data sets are provided in Table \ref{tab:stat}. The first one, which is referred to as $Tweet~Set{-}1$ in Table \ref{tab:stat}, is presented in \cite{Kucuk2014} and comprises 2,320 tweets with about 20K tokens. The second data set ($Tweet~Set{-}2$) includes about 5K tweets with about 50K tokens and is described in \cite{Eryigit2013}.

\begin{table*}[ht]
  \scriptsize
  \centering
  \caption{Initial NER Evaluation Results (Strict Metrics).}
    \begin{tabular}{|r|c|c|c|c|c|c|c|}
    \hline
    \multicolumn{1}{|l|}{  \textbf{Data Set}} & \textbf{Capitalization} & \textbf{Metric} & \textbf{Person} & \textbf{Location} & \textbf{Organization} & \textbf{Overall for PLOs} & \textbf{Overall for 7 Types} \\ \hline
    \multicolumn{1}{|c|}{\multirow{6}[11]{*}{\textbf{Tweet Set-1}}} & \multirow{3}[5]{*}{\textbf{ON}}

                               & \textbf{P (\%)} & 52.82 & 77.78 & 72.34 & 64.16 & 71.13 \\ \cline{3-8}
    \multicolumn{1}{|c|}{} &   & \textbf{R (\%)} & 32.82 & 49.65 & 28.22 & 36.53 & 42.53 \\ \cline{3-8}
    \multicolumn{1}{|c|}{} &   & \textbf{F (\%)} & \textbf{40.49} & \textbf{60.61} & \textbf{40.60} & \textbf{46.55} & \textbf{53.23} \\ \cline{2-8}

    \multicolumn{1}{|c|}{} & \multirow{3}[6]{*}{\textbf{OFF}}

                               & \textbf{P (\%)} & 36.73 & 71.72 & 58.70 & 49.29 & 56.21 \\ \cline{3-8}
    \multicolumn{1}{|c|}{} &   & \textbf{R (\%)} & 43.33 & 62.06 & 33.61 & 46.33 & 50.45 \\ \cline{3-8}
    \multicolumn{1}{|c|}{} &   & \textbf{F (\%)} & \textbf{39.76} & \textbf{66.54} & \textbf{42.74} & \textbf{47.76} & \textbf{53.18} \\ \hline

    \multicolumn{1}{|c|}{\multirow{6}[12]{*}{\textbf{Tweet Set-2}}} & \multirow{3}[6]{*}{\textbf{ON}}

                               & \textbf{P (\%)} & 55.79 & 58.68 & 72.06 & 58.86 & 65.62 \\ \cline{3-8}
    \multicolumn{1}{|c|}{} &   & \textbf{R (\%)} & 20.54 & 37.17 & 11.98 & 20.31 & 30.85 \\ \cline{3-8}
    \multicolumn{1}{|c|}{} &   & \textbf{F (\%)} & \textbf{30.03} & \textbf{45.51} & \textbf{20.55} & \textbf{30.19} & \textbf{41.97} \\ \cline{2-8}

    \multicolumn{1}{|c|}{} & \multirow{3}[6]{*}{\textbf{OFF}}

                               & \textbf{P (\%)} & 35.61 & 45.53 & 40.72 & 38.31 & 46.27 \\ \cline{3-8}
    \multicolumn{1}{|c|}{} &   & \textbf{R (\%)} & 38.37 & 61.26 & 16.63 & 35.08 & 42.40 \\ \cline{3-8}
    \multicolumn{1}{|c|}{} &   & \textbf{F (\%)} & \textbf{36.94} & \textbf{52.23} & \textbf{23.61} & \textbf{36.63} & \textbf{44.25} \\ \hline
    \end{tabular}
  \label{tab:eval1}
\end{table*}

\begin{table*}[ht]
  \scriptsize
  \centering
  \caption{Initial NER Evaluation Results (Partial Metrics).}
    \begin{tabular}{|r|c|c|c|c|c|c|c|}
    \hline
    \multicolumn{1}{|l|}{  \textbf{Data Set}} & \textbf{Capitalization} & \textbf{Metric} & \textbf{Person} & \textbf{Location} & \textbf{Organization} & \textbf{Overall for PLOs} & \textbf{Overall for 7 Types}\\ \hline
    \multicolumn{1}{|c|}{\multirow{6}[11]{*}{\textbf{Tweet Set-1}}} & \multirow{3}[5]{*}{\textbf{ON}}

                               & \textbf{P (\%)} & 65.33 & 86.05 & 88.37 & 75.98 & 80.74 \\ \cline{3-8}
    \multicolumn{1}{|c|}{} &   & \textbf{R (\%)} & 39.38 & 54.01 & 32.34 & 41.87 & 47.13 \\ \cline{3-8}
    \multicolumn{1}{|c|}{} &   & \textbf{F (\%)} & \textbf{49.14} & \textbf{66.37} & \textbf{47.35} & \textbf{53.99} & \textbf{59.52} \\ \cline{2-8}

    \multicolumn{1}{|c|}{} & \multirow{3}[6]{*}{\textbf{OFF}}

                               & \textbf{P (\%)} & 42.83 & 78.68 & 69.11 & 56.25 & 62.49 \\ \cline{3-8}
    \multicolumn{1}{|c|}{} &   & \textbf{R (\%)} & 50.92 & 67.71 & 38.00 & 52.55 & 55.72 \\ \cline{3-8}
    \multicolumn{1}{|c|}{} &   & \textbf{F (\%)} & \textbf{46.53} & \textbf{72.78} & \textbf{49.04} & \textbf{54.34} & \textbf{58.91} \\ \hline

    \multicolumn{1}{|c|}{\multirow{6}[12]{*}{\textbf{Tweet Set-2}}} & \multirow{3}[6]{*}{\textbf{ON}}

                               & \textbf{P (\%)} & 69.79 & 61.34 & 74.63 & 68.27 & 72.51 \\ \cline{3-8}
    \multicolumn{1}{|c|}{} &   & \textbf{R (\%)} & 24.28 & 38.62 & 12.25 & 22.65 & 33.31 \\ \cline{3-8}
    \multicolumn{1}{|c|}{} &   & \textbf{F (\%)} & \textbf{36.03} & \textbf{47.40} & \textbf{21.05} & \textbf{34.02} & \textbf{45.65} \\ \cline{2-8}

    \multicolumn{1}{|c|}{} & \multirow{3}[6]{*}{\textbf{OFF}}

                               & \textbf{P (\%)} & 41.82 & 48.41 & 41.99 & 43.21 & 50.91 \\ \cline{3-8}
    \multicolumn{1}{|c|}{} &   & \textbf{R (\%)} & 45.10 & 65.59 & 17.06 & 39.38 & 46.45 \\ \cline{3-8}
    \multicolumn{1}{|c|}{} &   & \textbf{F (\%)} & \textbf{43.40} & \textbf{55.71} & \textbf{24.26} & \textbf{41.21} & \textbf{48.58} \\ \hline

    \end{tabular}
  \label{tab:eval2}
\end{table*}

\subsection{Initial Experiments}\label{subsec:initial}
We have first evaluated the system's performance on the data sets without any extensions to the existing NER system. Table \ref{tab:eval1} presents these evaluation results using the commonly employed metrics of precision, recall, and balanced F-Measure, without giving any credit to partially extracted NEs. Table \ref{tab:eval2} displays those results with the same metrics this time giving credit to partial extractions with the constraint that the NE type within the system output and the answer key must be the same, where these metrics have been employed in studies like \cite{Maynard2001}.

The evaluation results in Table \ref{tab:eval1} and Table \ref{tab:eval2} are in line with the common finding reported in the literature that the NER systems for comparatively well-formed text types face considerable performance decreases when they are evaluated on tweets. This observation is usually attributed to the peculiarities of tweet texts such as common grammatical\slash spelling errors and deliberate contractions. With strict metrics, the system is reported to achieve an F-Measure rate of 78.7\%. When it is ported to tweets, the best overall F-Measure rates achieved are 53.23\% and 44.25\% on $Tweet~Set{-}1$ and $Tweet~Set{-}2$, respectively, while the corresponding best F-Measure rates for only PLOs are 47.76\% and 36.63\%, respectively, all with strict metrics. The difference between the results for PLOs and the overall results also confirms that the system recognizes temporal and numerical expressions (within its scope) with decent performance, compared to the recognition of PLOs.

The F-Measure rates obtained when partial extractions are also given credit are about 5\% higher than those obtained without giving any credit to partially extracted NEs. This increase is important due to pragmatic reasons as these partially extracted NEs can help conveniently filter tweet streams and retrieve relevant subsets of tweets in several application settings.

\subsection{NER Experiments with Rule\slash Resource Adaptations}\label{subsec:adapt}
Tweet texts possess the following peculiarities usually as opposed to other formal text types:

\begin{itemize}
    \item   Grammatical\slash spelling errors are common, like incorrectly writing proper names all in lowercase letters. A Turkish example illustrating a spelling error is the use of $geliyoooo$ instead of $geliyor$ (meaning $is~coming$).
    \item   Contracted word forms are commonly used instead of full forms, like referring to the football club called $Fenerbah$\emph{\c{c}}$e$ as $Fener$ only, where the latter contracted form is also homonymous to a common name in Turkish (meaning $lantern$).
    \item   For the particular case of Turkish tweets, non-accentuated characters (c, g, i, o, s, and u) are often utilized instead of the corresponding Turkish characters with diacritics (\c{c}, \u{g}, {\i}, {\"o}, \c{s}, and {\"u}). An example of this phenomenon is writing $cunku$ instead of the correct form, \emph{\c{c}\"u}$nk$\emph{\"u} (meaning $because$).
\end{itemize}

Considering the above features, in order to improve the initial NER performance on Turkish tweets, we have tested two adaptations of the rule-based NER system. The details of these adaptations and the corresponding evaluation results are presented in the following subsections.

\begin{table*}[ht]
  \scriptsize
  \centering
  \caption{NER Evaluation Results After Diacritics-Based Expansion of Resources (Strict Metrics).}
    \begin{tabular}{|r|c|c|c|c|c|c|c|}
    \hline
    \multicolumn{1}{|l|}{  \textbf{Data Set}} & \textbf{Capitalization} & \textbf{Metric} & \textbf{Person} & \textbf{Location} & \textbf{Organization} & \textbf{Overall for PLOs} & \textbf{Overall for 7 Types} \\ \hline
    \multicolumn{1}{|c|}{\multirow{6}[11]{*}{\textbf{Tweet Set-1}}} & \multirow{3}[5]{*}{\textbf{ON}}

                               & \textbf{P (\%)} & 53.00 & 78.80 & 73.20 & 64.89 & 71.95 \\ \cline{3-8}
    \multicolumn{1}{|c|}{} &   & \textbf{R (\%)} & 32.82 & 51.42 & 29.46 & 37.35 & 44.26 \\ \cline{3-8}
    \multicolumn{1}{|c|}{} &   & \textbf{F (\%)} & \textbf{40.54} & \textbf{62.23} & \textbf{42.01} & \textbf{47.41} & \textbf{54.81} \\ \cline{2-8}

    \multicolumn{1}{|c|}{} & \multirow{3}[6]{*}{\textbf{OFF}}

                               & \textbf{P (\%)} & 36.17 & 71.31 & 59.03 & 48.95 & 56.16 \\ \cline{3-8}
    \multicolumn{1}{|c|}{} &   & \textbf{R (\%)} & 43.76 & 63.48 & 35.27 & 47.35 & 52.35 \\ \cline{3-8}
    \multicolumn{1}{|c|}{} &   & \textbf{F (\%)} & \textbf{39.60} & \textbf{67.17} & \textbf{44.16} & \textbf{48.13} & \textbf{54.19} \\ \hline

    \multicolumn{1}{|c|}{\multirow{6}[12]{*}{\textbf{Tweet Set-2}}} & \multirow{3}[6]{*}{\textbf{ON}}

                               & \textbf{P (\%)} & 58.22 & 58.73 & 70.67 & 60.20 & 67.29 \\ \cline{3-8}
    \multicolumn{1}{|c|}{} &   & \textbf{R (\%)} & 22.87 & 38.74 & 12.96 & 22.13 & 34.89 \\ \cline{3-8}
    \multicolumn{1}{|c|}{} &   & \textbf{F (\%)} & \textbf{32.84} & \textbf{46.69} & \textbf{21.90} & \textbf{32.36} & \textbf{45.95} \\ \cline{2-8}

    \multicolumn{1}{|c|}{} & \multirow{3}[6]{*}{\textbf{OFF}}

                               & \textbf{P (\%)} & 36.80 & 44.61 & 32.43 & 37.61 & 46.24 \\ \cline{3-8}
    \multicolumn{1}{|c|}{} &   & \textbf{R (\%)} & 43.41 & 62.83 & 17.60 & 38.43 & 47.64 \\ \cline{3-8}
    \multicolumn{1}{|c|}{} &   & \textbf{F (\%)} & \textbf{39.83} & \textbf{52.17} & \textbf{22.82} & \textbf{38.01} & \textbf{46.93} \\ \hline

    \end{tabular}
  \label{tab:eval3}
\end{table*}

\begin{table*}[ht]
  \scriptsize
  \centering
  \caption{NER Evaluation Results After Diacritics-Based Expansion of Resources (Partial Metrics).}
    \begin{tabular}{|r|c|c|c|c|c|c|c|}
    \hline
    \multicolumn{1}{|l|}{  \textbf{Data Set}} & \textbf{Capitalization} & \textbf{Metric} & \textbf{Person} & \textbf{Location} & \textbf{Organization} & \textbf{Overall for PLOs} & \textbf{Overall for 7 Types}\\ \hline
    \multicolumn{1}{|c|}{\multirow{6}[11]{*}{\textbf{Tweet Set-1}}} & \multirow{3}[5]{*}{\textbf{ON}}

                               & \textbf{P (\%)} & 65.58 & 87.46 & 88.76 & 76.81 & 81.44 \\ \cline{3-8}
    \multicolumn{1}{|c|}{} &   & \textbf{R (\%)} & 39.38 & 56.12 & 33.62 & 42.80 & 48.98 \\ \cline{3-8}
    \multicolumn{1}{|c|}{} &   & \textbf{F (\%)} & \textbf{49.21} & \textbf{68.37} & \textbf{48.77} & \textbf{54.97} & \textbf{61.17} \\ \cline{2-8}

    \multicolumn{1}{|c|}{} & \multirow{3}[6]{*}{\textbf{OFF}}

                               & \textbf{P (\%)} & 42.21 & 79.17 & 69.00 & 56.02 & 62.49 \\ \cline{3-8}
    \multicolumn{1}{|c|}{} &   & \textbf{R (\%)} & 51.56 & 69.85 & 39.70 & 53.88 & 57.90 \\ \cline{3-8}
    \multicolumn{1}{|c|}{} &   & \textbf{F (\%)} & \textbf{46.42} & \textbf{74.22} & \textbf{50.40} & \textbf{54.93} & \textbf{60.11} \\ \hline

    \multicolumn{1}{|c|}{\multirow{6}[12]{*}{\textbf{Tweet Set-2}}} & \multirow{3}[6]{*}{\textbf{ON}}

                               & \textbf{P (\%)} & 71.48 & 61.29 & 72.97 & 69.07 & 73.68 \\ \cline{3-8}
    \multicolumn{1}{|c|}{} &   & \textbf{R (\%)} & 26.68 & 40.21 & 13.24 & 24.51 & 37.47 \\ \cline{3-8}
    \multicolumn{1}{|c|}{} &   & \textbf{F (\%)} & \textbf{38.86} & \textbf{48.56} & \textbf{22.41} & \textbf{36.18} & \textbf{49.67} \\ \cline{2-8}

    \multicolumn{1}{|c|}{} & \multirow{3}[6]{*}{\textbf{OFF}}

                               & \textbf{P (\%)} & 42.26 & 47.07 & 33.33 & 41.75 & 50.23 \\ \cline{3-8}
    \multicolumn{1}{|c|}{} &   & \textbf{R (\%)} & 50.14 & 66.76 & 18.04 & 42.65 & 51.72 \\ \cline{3-8}
    \multicolumn{1}{|c|}{} &   & \textbf{F (\%)} & \textbf{45.86} & \textbf{55.21} & \textbf{23.41} & \textbf{42.20} & \textbf{50.96} \\ \hline

    \end{tabular}
  \label{tab:eval4}
\end{table*}

\subsubsection{Relaxing the Capitalization Constraint of the System}\label{subsec:cap}
As proper capitalization of PLOs is usually lacking in tweets, we have evaluated the NER system with its capitalization feature turned off, so that the system considers all tokens (no matter whether their initial character is capitalized or not) as valid NE candidates. The initial evaluation results of the system with this setting are provided in Table \ref{tab:eval1} and Table \ref{tab:eval2} within the rows where the $Capitalization$ column has a corresponding $OFF$ value. The results for these two capitalization settings are also similarly provided in Tables \ref{tab:eval3}-\ref{tab:eval5} which present the evaluation results described in the upcoming sections.

The results in Table \ref{tab:eval1} and Table \ref{tab:eval2} demonstrate that relaxing the capitalization constraint (i.e., not using the capitalization clue) during the NER procedure on Turkish tweets consistently improves performance for PLOs on both data sets. The improvement obtained with this relaxation is more dramatic on $Tweet~Set{-}2$ and for this data set the overall results are accordingly better than those obtained when the capitalization clue is used. It should again be noted that the NER system uses a simplistic morphological analyzer to validate suffixes added at the ends of the NEs, thereby the system does not overgenerate with this new setting, although the precision rates decrease considerably in return to corresponding increases in the recall rates. To summarize, together with the fact that about 25.1\% of all PLOs within $Tweet~Set{-}1$ are lacking proper capitalization \cite{Kucuk2014}, these findings suggest that the ability to relax this capitalization constraint is a convenient feature of a practical NER system for Turkish tweets. An alternative feature would be to automatically correct the capitalization of NEs instead, as a pre-processing step.

\subsubsection{Diacritics-Based Expansion of the Lexical Resources}\label{subsec:diac}
In Turkish tweet texts, words including Turkish characters with diacritics are often, usually either erroneously or deliberately for pragmatic reasons such as to type faster, spelled with their non-diacritic equivalents, as pointed out above. Therefore, we expand the entries in the lexical resources of the NER system to include both diacritic and non-diacritic variants of these entries. For instance, the Turkish name of the island $Cyprus$, $K$\emph{{\i}}$br$\emph{{\i}}$s$, may appear in tweets as $K$\emph{{\i}}$bris$, $Kibr$\emph{{\i}}$s$, or $Kibris$, as well. As this example denotes, for each existing entry with $n$ such Turkish-specific characters, $2^n$ entries (including the original entry) are included in the ultimate expanded forms of the lexical resources, since each such character may be used as it is or may be replaced with its equivalent.

During this expansion stage, we have applied a filtering procedure over these newly considered $2^n-1$ entries to check whether they are homonymous to common names in Turkish. This filtering procedure basically checks whether an expansion candidate is within a list of unique, supposedly well-formed, Turkish words comprising about 1,140,208 items including inflected forms \cite{zemberek}, and if it is, then this candidate is discarded to avoid overgeneration during the actual NER procedure.

We have tested this new version of the system with expanded lexical resources and the corresponding evaluation results are provided in Table \ref{tab:eval3} and Table \ref{tab:eval4}, using the strict and partial evaluation metrics, respectively. Both strict and partial evaluation results denote that the performance of the system is improved after this diacritics-based expansion of the system resources. The best results are obtained when this expansion is combined with the relaxation of the capitalization constraint, for PLOs on $Tweet~Set{-}1$, and both for PLOs and all 7 NE types on $Tweet~Set{-}2$. Similar to the points made in the previous section, this diacritics-based expansion scheme stands as a promising feature of an ultimate NER system for Turkish tweets, also considering the fact that about 6.3\% of all NEs in $Tweet~Set{-}1$ are written in characters with missing diacritics. A plausible alternative to this feature would be to perform diacritics-based correction (or, normalization) as presented in studies like \cite{mihalcea2002diacritics} prior to the actual NER procedure. Similar approaches can be tested on tweets in other languages having common characters with diacritics.

\subsection{Tweet Normalization}\label{subsec:norm}
Tweet normalization has emerged as an important research problem \cite{Han2011}, the solutions to which can readily be used in systems for sentiment analysis and NER (as considered in studies such as \cite{Liu2012}), among others. In order to observe the effects of normalization on NER performance on Turkish tweets, we have first experimented with a simplistic tweet normalization scheme which aims at decreasing repeated characters in words, as repetition of characters in tweets is a frequent means to express stress. The scheme is outlined below:

\begin{enumerate}
    \item   In order to determine the list of valid Turkish words with consecutively repeated characters, we have employed the list of Turkish unique words \cite{zemberek}, that we have previously utilized during the diacritics-based resource expansion procedure in Section \ref{subsec:diac}. Within this list, 74,262 words (about 6.5\% of the list) turn out to have consecutively repeated characters.
    \item   Using this sublist as a reference resource, we have implemented the actual simplistic normalization scheme: if a word in a tweet has consecutively repeated character sequences and the word is not included within the aforementioned sublist, then all of these character sequences are contracted to single character instances. For instance, with this procedure, the token $zamaanlaaa$ is correctly replaced with $zamanla$ (meaning $with~time$) and $mirayyy$ is correctly replaced with $miray$ (a proper person name).
\end{enumerate}

The employment of the above normalization scheme prior to the actual NER procedure has led to slightly poorer results as some NEs which should not be normalized through this scheme are normalized instead. For instance, the city name \emph{\c{C}}$anakkale$ is changed to \emph{\c{C}}$anakale$ during the normalization procedure and it is missed by the subsequent NER procedure. Hence, we employ a three-phase pipelined NER approach where we first run the NER procedure on the input text, then employ the normalization scheme on the NER output, and finally run the NER procedure again on the normalization output, in order to avoid that the normalization step corrupts well-formed NEs that can readily be extracted by the system.

The performance of this ultimate NER pipeline, with the capitalization feature turned off during both of the actual NER phases, is evaluated only on $Tweet~Set{-}1$. Therefore, the performance evaluations of the first NER phase correspond to the previously presented results in the rows 4-6 of Table \ref{tab:eval1} and Table \ref{tab:eval2}, with strict and partial versions of the metrics, respectively.

Below we summarize our findings regarding the intermediate normalization procedure employed, based on its evaluation results. Although some of these findings are not directly relevant for the purposes of the NER procedure, we provide them for the completeness of the discussion on the normalization of Turkish tweets.

\begin{itemize}
    \item   Excluding the normalization cases which involve non-alphabetical characters only (like normalizing \texttt{>>>>>>} to \texttt{>}), those that result in a normalized form with a single alphabetical character (like normalizing \texttt{oooooo} to \texttt{o}), and those that involve emotion expressions (like normalizing \texttt{:DDDDD} to \texttt{:D}), the number of resulting instances considered for performance evaluation is 494.
    \item   The number of normalization instances in which an incorrect token is precisely converted into its corresponding valid form is 253, so, the precision of the overall normalization scheme is 51.21\%.
    \item   117 of the incorrect cases are due to the fact that the token that is considered for normalization is a valid but foreign token (such as normalizing $Harry$ to $Hary$, $jennifer$ to $jenifer$, $full$ to $ful$, and $tweet$ to $twet$). Hence, these cases account for a decrease of 23.68\% in the precision of the normalization scheme.
    \item   15 of the incorrect instances are due to the fact that Turkish characters with diacritics are not correctly used, hence they cannot be found within the reference sublist of valid Turkish words, and subsequently considered by the normalization procedure, although they could instead be subject to a diacritics-based normalization, as pointed out at the end of Section \ref{subsec:diac}. For instance, \emph{\c{s}}$iir$ (meaning $poem$) is incorrectly written as $siir$ in a tweet and since it, in this incorrect form, cannot be found on the reference sublist, it is erroneously changed to $sir$. There are also other incorrect instances in which superfluous characters are correctly removed with the normalization procedure, yet the resulting token is still not in its correct form as a subsequent diacritics-based correction is required. Though they are not considerably frequent (as we only consider here tokens with consecutively repeated characters), these instances serve to confirm that the restoration of diacritics should be considered along with other forms of normalization.
    \item   Some other frequent errors made by the normalization scheme are due to incorrect tokenization as whitespaces to separate tokens can be missing due to writing errors or the tendency to write some phrases hashtag-like. An example case is incorrectly writing the adverb, $demek~ki$ (meaning $so$ or $that~means$), as $demekki$ in a tweet, which in turn is erroneously changed to $demeki$ during normalization. This token, $demekki$, should not be considered within this type of normalization at all, although it needs processing to be transformed into its correct form, $demek~ki$.
\end{itemize}

To summarize, the normalization scheme can be enhanced considering the above points, where proper treatment of non-Turkish tokens and the consideration of diacritics-based issues stand as the most promising directions of improvement. Other more elaborate ways of normalizing tweets, as presented in studies such as \cite{Han2011}, should also be tested together with the NER procedure, to observe their ultimate contribution. Along the way, a normalization dictionary for Turkish can be compiled, following studies like \cite{Han2012}.

\begin{table*}
  \scriptsize
  \centering
  \caption{Evaluation Results of the NER Pipeline with Normalization, on $Tweet~Set{-}1$.}
    \begin{tabular}{|l|c|c|c|c|c|c|}
    \hline
    \textbf{Metric Type} & \textbf{Metric} & \textbf{Person} & \textbf{Location} & \textbf{Organization} & \textbf{Overall for PLOs} & \textbf{Overall for 7 Types} \\ \hline
    \multirow{3}[5]{*}{\textbf{Strict}}
          & \textbf{P (\%)} & 36.45 & 71.72 & 58.99 & 48.94 & 55.91 \\ \cline{2-7}
          & \textbf{R (\%)} & 44.42 & 62.06 & 34.02 & 46.94 & 51.20 \\ \cline{2-7}
          & \textbf{F (\%)} & \textbf{40.04} & \textbf{66.54} & \textbf{43.16} & \textbf{47.92} & \textbf{53.45} \\ \hline
    \multirow{3}[6]{*}{\textbf{Partial}}
          & \textbf{P (\%)} & 42.32 & 78.68 & 69.35 & 55.73 & 62.04 \\ \cline{2-7}
          & \textbf{R (\%)} & 52.07 & 67.71 & 38.43 & 53.18 & 56.48 \\ \cline{2-7}
          & \textbf{F (\%)} & \textbf{46.69} & \textbf{72.78} & \textbf{49.45} & \textbf{54.43} & \textbf{59.13} \\ \hline
    \end{tabular}
  \label{tab:eval5}
\end{table*}

The evaluation results of the ultimate three-phase NER pipeline are provided in Table \ref{tab:eval5}, with the systems's capitalization feature turned off in both NER phases. Within the first three rows, the results with the strict evaluation metrics are displayed while the last three rows present those results obtained with the partial versions. When we examine the individual NER results after the incorporation of normalization scheme in details, we observe that there are cases where incorrectly normalizing some common names or slang\slash contracted words leads to them being extracted as NEs during the second NER phase. In order to prevent such false positives, the ways of improving the normalization procedure discussed above can be implemented and thereby less errors will be propagated into the second NER phase.

Though the overall results in Table \ref{tab:eval5} are slightly better than their counterparts when normalization is not employed, we cannot derive sound conclusions about the contribution of this normalization scheme to the overall NER procedure. The slight improvement is also an expected result as the size of the test data set is quite small and the number of NEs to be recognized after this type of normalization is already limited since only about 1\% of all PLOs in $Tweet~Set{-}1$ have incorrectly repeated consecutive characters. Yet, the results are still promising in that with a more elaborate normalization procedure evaluated on larger corpora, more dramatic increases in the NER performance can be obtained on Turkish tweets.

\section{Future Work}\label{sec:future}
Directions of future work based on the current study include the following:
\begin{itemize}
    \item   Following the points made throughout Section \ref{sec:tests}, several normalization schemes also involving case and diacritics restoration can be implemented and incorporated into the NER procedure on tweets.
    \item   Since tweet texts are short and informal, they often lack contextual clues needed to perform an efficient NER procedure. Additionally, there is a tendency to mention new and popular NEs in tweets which might be missed by a NER system with static lexical resources. Hence, extending the lexical resources of the NER system with contemporary up-to-date NEs automatically obtained from Turkish news articles can be considered. For this purpose, we can readily employ resources like JRC-Names \cite{Steinberger2011jrc}, a publicly available continuously-updated NE and name variant dictionary, as a source of up-to-date NEs in Turkish.
\end{itemize}

\section{Conclusion}\label{sec:conc}
In this study, we target the problem of named entity recognition on Turkish tweets. We have carried out experiments starting with a rule-based recognition system and gradually extended it in two directions: adapting the rules\slash resources of the system and introducing a tweet normalization scheme into the recognition procedure. Thereby, we present our findings on named entity recognition on Turkish tweets in addition to those on the normalization of Turkish tweets. Based on these findings, we outline some desirable features of a named entity recognition system tailored to Turkish tweets. Future work includes the employment and testing of more elaborate tweet normalization procedures along the way, on larger tweet data sets, in addition to evaluating the system after its resources are automatically extended with dictionaries of up-to-date named entities.

\section*{Acknowledgments}
This study is supported in part by a postdoctoral research grant from T\"UB\.ITAK.

\bibliographystyle{acl}
\bibliography{twinertr}

\end{document}